\documentclass{article}
\usepackage{spconf,amsmath,graphicx}
\usepackage{amssymb}
\usepackage{hyperref}
\usepackage[dvipsnames]{xcolor}
\usepackage[ruled,vlined,linesnumbered]{algorithm2e}

\makeatletter
\newcommand{\removelatexerror}{\let\@latex@error\@gobble}
\makeatother

\title{MULTI-OBJECT TRACKING BY ITERATIVELY ASSOCIATING DETECTIONS WITH UNIFORM APPEARANCE FOR TRAWL-BASED FISHING BYCATCH MONITORING}
\name{\begin{tabular}{@{}c@{}}
Cheng-Yen Yang$^{1}$ \quad 
Alan Yu Shyang Tan$^{2}$ \quad 
Melanie J. Underwood$^{2}$ \quad
Charlotte Bodie$^{2}$\\ 
Zhongyu Jiang$^{1}$ \quad 
Steve George$^{2}$ \quad 
Karl Warr$^{3}$ \quad
Jenq-Neng Hwang$^{1}$ \quad
Emma Jones$^{2}$
\end{tabular}}
\address{$^{1}$ Department of Electrical $\&$ Computer Engineering, University of Washington, United States \\$^{2}$ National Institute of Water and Atmospheric Research, New Zealand \  $^{3}$ BetterFishing Ltd., New Zealand}

\begin{document}

\maketitle

\begin{abstract}
The aim of in-trawl catch monitoring for use in fishing operations is to detect, track and classify fish targets in real-time from video footage. Information gathered could be used to release unwanted bycatch in real-time. However, traditional multi-object tracking (MOT) methods have limitations, as they are developed for tracking vehicles or pedestrians with linear motions and diverse appearances, which are different from the scenarios such as livestock monitoring. Therefore, we propose a novel MOT method, built upon an existing observation-centric tracking algorithm, by adopting a new iterative association step to significantly boost the performance of tracking targets with a uniform appearance. The iterative association module is designed as an extendable component that can be merged into most existing tracking methods. Our method offers improved performance in tracking targets with uniform appearance and outperforms state-of-the-art techniques on our underwater fish datasets as well as the MOT17 dataset, without increasing latency nor sacrificing accuracy as measured by HOTA, MOTA, and IDF1 performance metrics.
\end{abstract}


\section{Introduction}
\label{sec:intro}

Bycatch \cite{bycatch} refers to the incidental catching of marine species that are not targeted. Some bycatch is valuable and utilized.  Unwanted bycatch is an undesirable catch that cannot be sold or is not allowed to be caught either due to size regulations or species restrictions, in commercial fisheries. Unwanted bycatch may include threatened and protected marine species, often resulting in unintentional harm and death. To combat this issue, both regulators and businesses in the global fishing industry are dedicated to reducing the occurrences of bycatch through innovative and technologically advanced methods.

\begin{figure}[t]
    \centering
    \includegraphics[width=.5\textwidth]{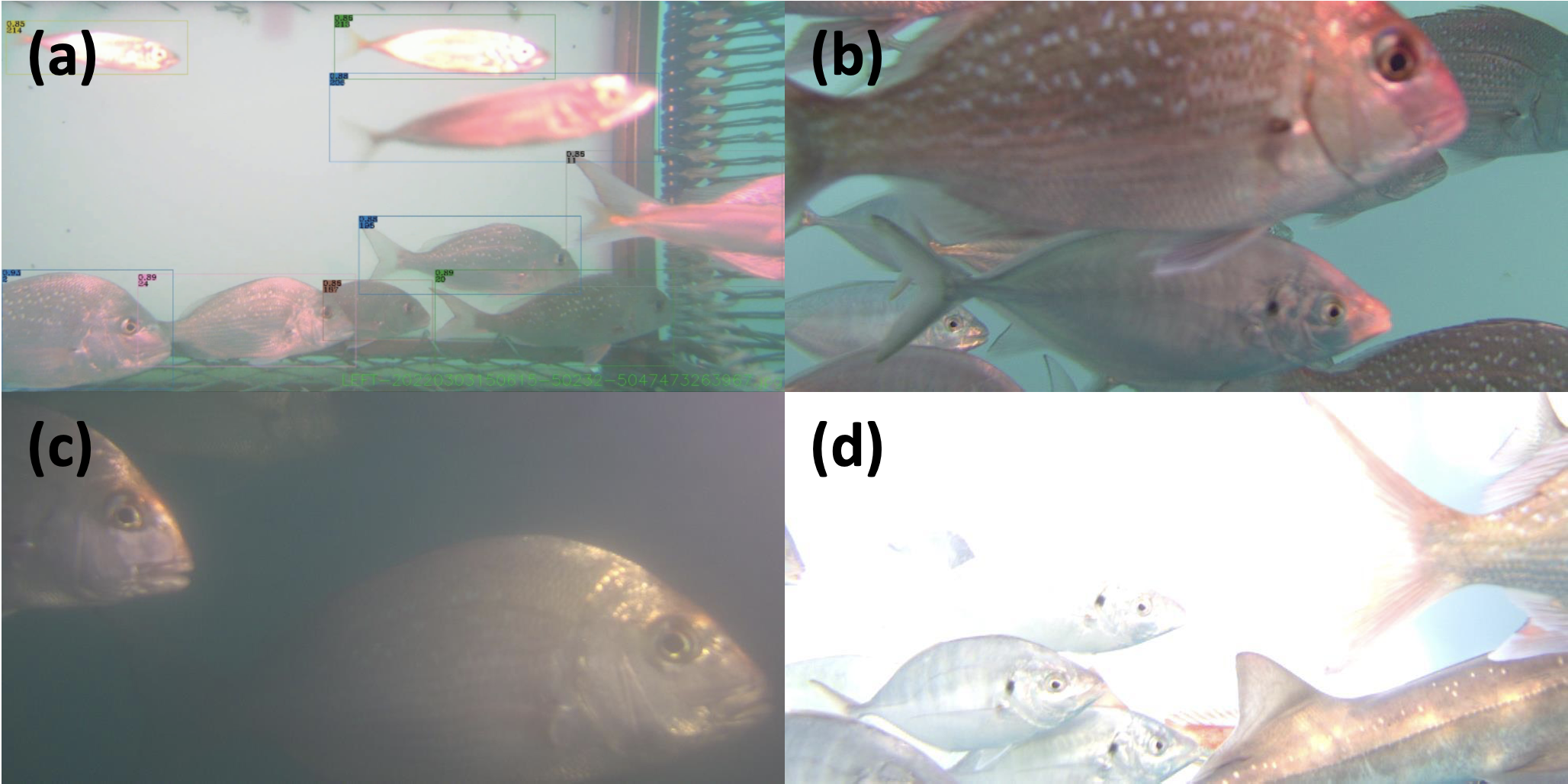}
    \vspace{-1em}
    \caption{Samples from our NIWA underwater fish dataset. (a) Visualization of our tracking on testing sequence, (b) heavily occluded targets, (c-d) highly varying lighting conditions. }
    \vspace{-1.5em}
\end{figure}

To minimize bycatches and discards in commercial fishing, traditional methods like altering the mesh size and shape of nets and incorporating escape panels into fishing gear have been explored. These strategies depend on size and shape differences between the target and non-target species to adequately separate the catch \cite{steven2007} are not always very effective. Challenges that need to be addressed for a robust solution for real-world viable selective fishing include:

\begin{itemize}
    \item High precision in detecting, recognizing, and tracking individual fish and their movements to allow the timely release of non-target species.
    \vspace{-1em}
    \item Fast inference speed to allow a reliable semi-automated release system to be deployed in commercial fisheries.
    \vspace{-1em}
    \item Overall systems need to be robust to challenging and highly varying environmental conditions, such as low lighting, low clarity, and data noise from occlusions from schools of fishes.
\end{itemize}

\let\thefootnote\relax\footnotetext{*This work was funded by the New Zealand Ministry of Business, Innovation and Employment’s Endeavour Science Investment Fund under contract CO1X1913: “Underwater selection tools for environmentally and economically sustainable fishing”, together with support from Industry partner BetterFishing Ltd.}

Previous studies on bycatch have looked into using modern deep learning models, which have demonstrated success in object detections, but we believe that the high similarity in appearance between individual fish, both inter-class and intra-class, is hindering their tracking-by-detection performance. To address this, we have developed a multi-object tracking (MOT) algorithm specifically designed for target objects having uniform appearances and complex movement patterns.

The remaining sections of the paper are organized as follows. Recent related works in multiple-object tracking (MOT) and trawl-based bycatch methods are given in Sec \ref{sec:related_work}. Then the methodology of our proposed tracking algorithm will be described in Sec \ref{sec:methods}. Finally, the details of the dataset and experimental results will be discussed in Sec \ref{sec:exp} followed by the conclusion in Sec \ref{sec:conclusion}.

\section{Related Works}
\label{sec:related_work}

\subsection{Trawl-based Fishing Applications}
Earlier pattern recognition works \cite{fish1, fish2} have focused on recognizing fish species by detecting or extracting some specified features based on image processing or computer vision techniques while more recent deep learning-based approaches \cite{deepfish1, fishcount, fishsize, fishsustain} use detection models to detect and classify fish in unconstrained underwater videos showing that a deep learning model trained to estimate fish abundance can outperform humans with satisfactory latency and accuracy. Furthermore, trawl-specific systems such as the CamTrawl \cite{camtrawl} and the DeepVision stereo camera system \cite{DeepVision} use a suite of stereo and digital cameras to capture videos and images of marine species. Other in-trawl systems \cite{intrawl1, intrawl2} utilize captured videos with different methods, such as stereo triangulation and multi-target tracking algorithms, for detecting fish and estimating lengths \cite{fishsize}, tracking and counting fish \cite{fishcount},  as well as identifying different fish species \cite{fishsustain}.

\subsection{Multiple-Object Tracking}

Multiple-Object Tracking (MOT), which involves identifying and keeping track of multiple objects within a video sequence, is a challenging problem in computer vision. One approach to MOT is known as tracking-by-detection, where object detections obtained from detectors \cite{He2017maskrcnn, wang2019jde, yolox2021} are utilized to track the same objects in subsequent frames. The tracking process can be accomplished through techniques such as Kalman filters \cite{Kalman1960} or deep learning models, which allow accurate prediction of the object locations in the next frame. Various data association methods such as SORT \cite{bewley2016sort}, DeepSORT\cite{wojke2017deepsort}, FairMOT\cite{zhang2021fairmot}, MeMOT\cite{Cai2022memot} utilized the location and motion cues or appearance features in different manners to achieve robust tracking results. BYTE\cite{zhang2022bytetrack} proposes a generic association method that associates almost every detection box instead of only the high score ones. OC-SORT\cite{cao2022observation} designs a pure motion model-based tracker which can improve tracking accuracy in crowded scenes. Recently, tracking objects with similar appearance e.g., group dancing \cite{sun2022dance}, sports tracking \cite{Huang_2023_WACV}, livestock monitoring, has attracted attention in which the previous re-ID procedure may fail due to non-discriminative features.

\section{Methods}
\label{sec:methods}

\subsection{Tracking}

\noindent \textbf{Iterative Association.} The idea from BYTE\cite{zhang2022bytetrack} of keeping not only the high confidence score detections but also the lower ones is an extremely crucial key to the success of our targeted scenes, where the fish are often being occluded by each other with only a partial portion of them being visible. Unlike the original implementation that splits the bounding boxes into two parts by a pre-determined score threshold, we design and modify this step of the algorithm into any arbitrary number $n$ of parts by a score thresholding set. The intuition behind this design is an observation of that for long-tailed distribution or even novelty class in the video data. We observe that the confidence score is not only due to occlusion or motion blurs as mentioned in \cite{zhang2022bytetrack} but also due to the few-shot or zero-shot detections that we need to deal with.

Our proposed tracking algorithm starts with an image frame $f_t$ where we retrieve the detection bounding boxes in terms of a series of $(x,y,w,h)$ vectors, ie., $X_{all}=\{x_1, x_2, \cdots\}$, $x_i \in \mathbb{R}^4$, and detection confidence score $x_i.conf \in [0,1]$ from a detector $\mathcal{D}$. These detections are assigned to different groups of $X$ according to their confidence scores: 

\vspace{-1em}
\begin{equation}
    X_m = \{x_i | c_m \leq x_i.conf \leq c_{m-1}, 0 \leq i \leq N\},
\end{equation}

\noindent which is used for a later stage of iterative association. Note that $c_0$ is typically a very low threshold and $c_{n}$ is the upper bound of confidence score which is $1$.

To clearly illustrate the idea, without loss of generality, we can use $n=3$ as an example, the $X_{all}$ are partitioned into three confidence-score groups partitions $\{X_1, X_2, X_3\}$ or $\{X_{high}, X_{med}, X_{low}\}$, given the confidence thresholds $c_1$ and $c_2$ as hyper-parameters. The $X_{high}$ is a set of detections containing all the bounding boxes with a confidence score higher than $c_2$, and the ones with a confidence score between $c_2$ and $c_1$ are assigned to $X_{med}$. The remaining ones belong to $X_{low}$. As for $n=2$, the detections are divided into two groups, and the tracking procedure will be identical to \cite{zhang2022bytetrack}. For $n=1$, the tracking algorithm will only associates the detections that are above the threshold, just like standard methods \cite{bewley2016sort, wojke2017deepsort, zhang2021fairmot, cao2022observation}.

We then iteratively associate the detections with the tracks. If the track is matched with a detection after the linear assignment using Hungrian algorithm, we remove such detection and track respectively. Then the unmatched tracks are associated with the next batch of detections (lower confidence scores as we literately progress).

\noindent \textbf{Cost Functions.} Following the assumption that detections with lower confidence scores often contain severe occlusion making the appearance feature unsatisfactory to be used as the associated cost. The cost function for the very first iteration is a mixture of RE-ID and IoU while the latter iterations use the IoU solely:

\begin{equation}
    C(X, \mathcal{T}) = \lambda_{IoU}C_{IoU}(X, Z_{\mathcal{T}}) + \lambda_{ReId} C_{ReId}(X, X_{\mathcal{T}})
\end{equation}

\noindent where $X \in \mathbb{R}^{N_{x}x5}$ stands for observations  and $Z \in \mathbb{R}^{N_{\mathbb{T}\times8}}$ stands for predicted states. For the ReID part, we do not specifically train an additional Re-ID feature extractor but use the feature output of the detector instead, i.e., the Yolo-X detector.

\noindent \textbf{Observation-Centric Processing.} In addition, we adopt the Observation-centric Online Smoothing (OOS) and Observation-Centric Recovery (OCR) modules from the OC-SORT \cite{cao2022observation} as these components are proved to be essential since the recovery of lost tracks and reduction of errors accumulated by linear motion models during the lost period is crucial in long time-span MOT. Observation-centric online smoothing can be seen as a dynamic and adaptable approach that balances the trade-off between tracking accuracy and computational efficiency, especially when dealing with large-scale and complex scenarios. When a track remains unassigned after the standard association phase, we turn to heuristics to associate its final observation with new observations in the incoming time step. This localized approach can effectively handle cases where an object may temporarily stop or become obscured. 

\noindent \textbf{Track Initialization and Removal.} Finally, we initialize a new track for each unmatched detection if the confidence is greater than some predefined threshold and also remove the old tracks if they are being unmatched for more than the predefined expiration time. With the tracks being successfully matched and tracked, we can then move on to the consecutive timestamp of the given sequence.

\subsection{Track-based Inferencing}
Due to the highly-occluded nature of our underwater fish dataset, the detections often contain only part of the whole fish boy, e.g., tail or head only. Therefore, inter-class misclassification is often observed during image-based classification. Even though our system, detector, and classifier, utilize image-based training, where the loss is calculated for each individual input image. But for the purpose of properly releasing non-targeted species, we implemented track-based inference schemes \cite{Mei2020VideobasedHS} to compare with image-based results. 

For each detection $x^t_{i}$ from a given track $t$, the logits $v(x^t_{i})$ are generated after the cropped region is sent to a separate classifier. The method refers as \textit{majority-voting} is a common technique to model ensembling, where two slightly derived schemes are implemented where the predictions in the first one are decided by the majority of instance-level prediction while the latter one is simply finding the arg max of the sum of the logits across the same track. We argue that by establishing track-based inferencing, it can provide an additional assessment of the actual performance on top of existing metrics and offers a better evaluation for the targeted fishery science communities.

\newcommand{\myalgorithm}{%
\begingroup
\removelatexerror
\begin{algorithm}[H]
 \footnotesize
 \caption{Pseudo-code of Iterative Association Tracking}
\SetAlgoLined
\KwData{Detector $\mathcal{D}$, Kalman Filter $KF$, cost functions, conf thresholds $[c_0=1, c_1, ...,c_n, c_{n+1}=0]$}
\KwResult{Tracks $\mathcal{T}$}
 Initialization: $\mathcal{T}\leftarrow \{\}$, $X_1, X_2, ..., X_n \leftarrow \{\}$\;
 \For{frame $f_t$, $t \leftarrow 1:T$}{
    \texttt{\\}
    \textcolor{Green}{// Step 1: Predict bboxes (with confidence scores) and tracks} \\
    $X \leftarrow \mathcal{D}(f_t)$ \;
    \For{$t$ in $\mathcal{T}$}{
        $t \leftarrow KF(t)$
    }
    \For{$m \leftarrow 1:n$}{
        $X_m \leftarrow \{x_i | c_m \leq x_i.conf \leq c_{m-1}, 0 \leq i \leq n\}$ 
    }
    \texttt{\\}
    \textcolor{Green}{// Step 2: Iteratively associate bboxes with tracks} \\
    $\mathcal{T}^{\text{unmatched}} \leftarrow \mathcal{T}, \mathcal{T}^{\text{matched}}\leftarrow \{\}, X^{\text{unmatched}}\leftarrow \{\}$\\
    \For {$m \leftarrow 1:n$}{
        Associate $\mathcal{T}^{\text{unmatched}}$ and $X_m$ by predefined cost function \\
        $\mathcal{T}^{\text{unmatched}} \leftarrow $ unmatched tracks from current iteration \\
        $\mathcal{T}^{\text{matched}} \leftarrow $ $\mathcal{T}^{\text{matched}} \cup $ matched tracks from current iteration \\
        $X^{\text{unmatched}}_m \leftarrow $ unmatched bboxes from current iteration \\
    }
    $X^{\text{unmatched}} \leftarrow \{X^{\text{unmatched}}_1, ..., X^{\text{unmatched}}_n\}$ \\

    \texttt{\\}
    \textcolor{Green}{// Step 3: Recover lost tracks using Observation-Centric Recovery} \\
    Associate $X^{\mathcal{T}^{\text{unmatched}}}_{\text{previous}}$ and $X^{\text{unmatched}}$ by predefined cost function \\
    $\mathcal{T}^{\text{unmatched}} \leftarrow $ unmatched tracks from current iteration \\
    $\mathcal{T}^{\text{matched}} \leftarrow $ $\mathcal{T}^{\text{matched}} \cup $ matched tracks from current step \\
    $X^{\text{unmatched}} \leftarrow $ unmatched bboxes from current iteration \\
    
    \texttt{\\}
    \textcolor{Green}{// Step 4: Perform Observation-Centric Online Smoothing} \\
    \For{$t$ in $\mathcal{T}^{\text{matched}}$}{
        $t \leftarrow \text{OOS}(t)$\\
    }

    \texttt{\\}
    \textcolor{Green}{// Step 5: Create new tracks and delete old tracks} \\
    \For{$x$ in $X_{\text{unmatched}}$}{
        $\mathcal{T}^{\text{new}} \leftarrow $ initial new tracks with unmatched detections
    }
    \For{x in $\mathcal{T}_{\text{unmatched}}$}{
        $\mathcal{T}^{\text{unmatched}} \leftarrow $ preserve old tracks within expiration threshold
    }
    $\mathcal{T} \leftarrow \{\mathcal{T}^{\text{matched}} , \mathcal{T}^{\text{new}} , \mathcal{T}^{\text{unmatched}} \}$\\
    
 }
\end{algorithm}
\endgroup}

\myalgorithm

\section{Experiments}
\label{sec:exp}

\subsection{Settings}

\noindent\textbf{Dataset.} We present a fine-grained underwater fish dataset, collected from Napier, Hawkes Bay region of New Zealand to specifically serves as the training resource and testing benchmarks for the automated image analysis of underwater fish analysis for trawl-based bycatch monitoring. Our datasets include 5 distinct data acquisition trips in the 13- months time span, which allows us to capture diversified intra-class and inter-class samples. We manually annotated 2000 images with a total of 5600 instances along with 21 specie labels by the experts. Furthermore, we selected 8 sequences from the testing dataset to conduct instance-level annotations in order to verify the performance of our MOT method. The training and testing sets are split based on the trips. The uncontrollable environmental conditions, such as low lighting and poor visibility, are presented along with the self-occlusion scenes. 
We also used the MOT17 \cite{MOT16} half-val to evaluate our proposed method for the generalization abilities of MOT.

\noindent\textbf{Evaluation Metrics.} We use the CLEAR metrics \cite{CLEARMOT}, including MOTA, IDF1, FP, FN, IDs, and HOTA to evaluate different aspects of the tracking performance. MOTA is computed based on false positive (FP), false negative (FN), and identity switch (IDs) while IDF1 evaluates the identity preservation ability and focuses more on the association's performance. In addition to tracking-related metrics, we also report overall top-1 and top-3 accuracy of the fish species classification results on our NIWA dataset.

\noindent\textbf{Implementation Details.} We adopt YOLO-x \cite{yolox2021} as our detector, which is trained on our self-collected NIWA dataset from a COCO-pretrained model as initialized weights. The detector and tracking are implemented on the Open MMLab \cite{mmtrack2020} framework, an open-source project based on PyTorch. The model is trained and tested on a single NVIDIA Quadro GV100 GPU.

\subsection{Experimental Results}

\noindent\textbf{Tracking on NIWA dataset.} In Table \ref{table:niwa}, we report the results of  tracking performance on NIWA dataset along with other benchmarks \cite{wojke2017deepsort, zhang2022bytetrack, cao2022observation} using  exactly the same Yolo-x detector. To provide a fair comparison between our proposed method and existing works, the tracking hyperparameters such as tracks initialize score, maximum tracklet lifetime and the first confidence thresholding of DeepSORT, BYTE and OC-SORT and our works are identical. The only additional parameters we have are the additional thresholding values in our system. Interpolations or other standard post-processing are not considered in this experiment as well.

\begin{table}[h]
\centering
\begin{tabular}{l|ccccc}
\hline
Method     & MOTA$\uparrow$ & IDF1$\uparrow$ & FP$\downarrow$  & FN$\downarrow$  & IDs$\downarrow$ \\ \hline
DeepSORT \cite{wojke2017deepsort}   & 64.9 & 55.3 & 198   & 255   & 175 \\
BYTE \cite{zhang2022bytetrack}       & 81.5 & 78.4 & 125 & 231 & 139 \\
OC-SORT \cite{cao2022observation}   & 79.8 & 80.4 & 133   & 208   & 124 \\ \hline
Ours (n=2) & 82.0 & \textbf{81.9} & 120   & 235   & 115 \\
Ours (n=3) & \textbf{82.5} & 80.7 & 105 & 199 & \textbf{110} \\ \hline
\end{tabular}
\caption{Comparison of the state-of-the-art methods using the same detections on NIWA test set. Note that for fair comparison, we used the same thresholding in BYTE and Ours($n=2$) experiments while tuning Ours($n=3$) separately.}
\label{table:niwa}
\end{table}

\vspace{.5em}

Our method provides an improvement on all MOTA, IDF1, FP, FN and IDS on our collected NIWA dataset as the figure also shows that our method can constantly outperform or give comparable result despite the decision on thresholding value selection.


\noindent\textbf{Species Identification on NIWA dataset.} We also report the accuracy of species identification in terms of track-based and image-based predictions using the same pre-trained classifier \cite{xie2017resnext}. As seen from Table \ref{table:niwa_cls}, the track-based method outperforms the image-based prediction as we observed that a great number of misclassifications in the image-based scheme resulting from the partially visible body part of the targets either due to occlusion or entering (exiting) the camera. These mistakes can further be corrected if we apply the tracked-based method since the target can appear in the view for a certain period of time, therefore by applying proper tracking, we can greatly improve the performance of our bycatch application. 

\begin{table}[h]
\centering
\begin{tabular}{l|cc}
\hline
Method     & Top-1 Acc. $\uparrow$ & Top-3 Acc. $\uparrow$  \\ \hline
Image-based    & 62.3 & 85.3 \\ 
Track-based   & \textbf{75.8} & \textbf{89.5} \\ \hline
\end{tabular}
\caption{Top-1 and Top-3 overall accuracy for species identification on the NIWA test set.}
\label{table:niwa_cls}
\end{table}

\noindent\textbf{Tracking on MOT17 Benchmark.} We also conduct our experiments on MOT benchmark like MOT17 in Table \ref{table:mot17}, as a generalizable method, our $n=2$ version can slightly outperform both OC-SORT and BYTE on the half-val benchmark without tuning the tracking parameters. Proving the overall generalization ability of our method. The results on \cite{wojke2017deepsort, zhang2022bytetrack, cao2022observation} we compared to were directly obtained from the Open MMLab mmtracking repository.

\begin{table}[h]
\centering
\begin{tabular}{l|cccc}
\hline
Method     & HOTA$\uparrow$ & MOTA$\uparrow$ & IDF1$\uparrow$  & IDs$\downarrow$ \\ \hline
DeepSORT \cite{wojke2017deepsort}  & - & 65.1 & 65.3    & 1199 \\
BYTE \cite{zhang2022bytetrack}      & 67.7 & 78.6 & 79.2  & 666 \\
OC-SORT \cite{cao2022observation}    & 67.5 & 77.8 & 78.4      & 825 \\ \hline
Ours (n=2) & \textbf{68.1} & \textbf{78.9} & \textbf{79.7}    & \textbf{646} \\ \hline
\end{tabular}
\caption{Comparison of the state-of-the-art methods using the same detections on MOT17 half-val using private detections.}
\label{table:mot17}
\end{table}

\vspace{-1em}
\section{Conclusion}
\label{sec:conclusion}

We proposed a novel MOT method that addresses the limitations of previous tracking methods by adopting a new iterative association step and combining it with the strengths of observation-centric algorithms along with a video-based underwater fish dataset that provides an alternative benchmark for tracking in uniform appearances. The proposed method offers improved performance in tracking targets with uniform appearance and outperforms existing techniques without sacrificing accuracy or increasing latency which is suitable to work in commercial fisheries applications.

\newpage

\vfill\pagebreak

\bibliographystyle{IEEEbib}

\small\bibliography{strings,refs}

\end{document}